\pdfoutput=1

\documentclass[11pt]{article}
\usepackage{algorithmic,algorithm}

\usepackage[]{EMNLP2022}
\usepackage{makecell}

\usepackage{times}
\usepackage{latexsym}
\usepackage{graphicx}
\usepackage{booktabs}
\usepackage[T1]{fontenc}
\usepackage{multirow}
\usepackage{latexsym}
\usepackage{multirow}
\usepackage[T1]{fontenc}

\usepackage[utf8]{inputenc}

\usepackage{microtype}
\usepackage{makecell}

\usepackage[utf8]{inputenc}

\usepackage{microtype}
\usepackage{amsfonts}

\usepackage{inconsolata}

\title{AdaPrompt: Adaptive Model Training for Prompt-based NLP}

\author{
 Yulong Chen$^{\spadesuit \heartsuit}$\thanks{~Yulong Chen completed this work during his internship at Microsoft.}\hspace{0.5mm}, 
 Yang Liu$^\clubsuit$\hspace{0.5mm}, 
 Li Dong$^\clubsuit$\hspace{0.5mm}, 
 Shuohang Wang$^\clubsuit$\hspace{0.5mm},\\
 \textbf{Chenguang Zhu$^\clubsuit$\hspace{0.5mm},
 Michael Zeng$^\clubsuit$\hspace{0.5mm}, 
 Yue Zhang$^{\heartsuit \diamondsuit}$\hspace{0.2mm}\hspace{1.5mm}} \\
 $^\spadesuit$ Zhejiang University \quad
 $^\heartsuit$ Westlake University\\
 $^\clubsuit$ Microsoft   Research \quad
 $^\diamondsuit$ Westlake Institute for Advanced Study\\
 \textit{\href{mailto:yulongchen1010@gmail.com}{yulongchen1010@gmail.com}} \quad\textit{\href{mailto:yaliu10@microsoft.com}{yaliu10@microsoft.com}}
 \quad\textit{\href{mailto:yue.zhang@wias.org.cn}{yue.zhang@wias.org.cn}}
}

\begin{document}
\maketitle
\begin{abstract}
Prompt-based learning, with its capability to tackle zero-shot and few-shot NLP tasks, has gained much attention in community.
The main idea is to bridge the gap between NLP downstream tasks and language modeling (LM), by mapping these tasks into natural language prompts, which are then filled by pretrained language models (PLMs).
However, for prompt learning, there are still two salient gaps between NLP tasks and pretraining.
First, prompt information is not necessarily sufficiently present during LM pretraining. 
Second, task-specific data are not necessarily well represented during pretraining. 
We address these two issues by proposing AdaPrompt, adaptively retrieving external data for continual pretraining of PLMs by making use of both task and prompt characteristics. 
In addition, we make use of knowledge in Natural Language Inference models for deriving adaptive verbalizers.
Experimental results on five NLP benchmarks show that AdaPrompt can improve over standard PLMs in few-shot settings. 
In addition, in zero-shot settings, our method outperforms standard prompt-based methods by up to 26.35\% relative error reduction.
\end{abstract}

\section{Introduction}\label{intro}
\begin{figure}
    \centering
    \includegraphics[width=1\columnwidth]{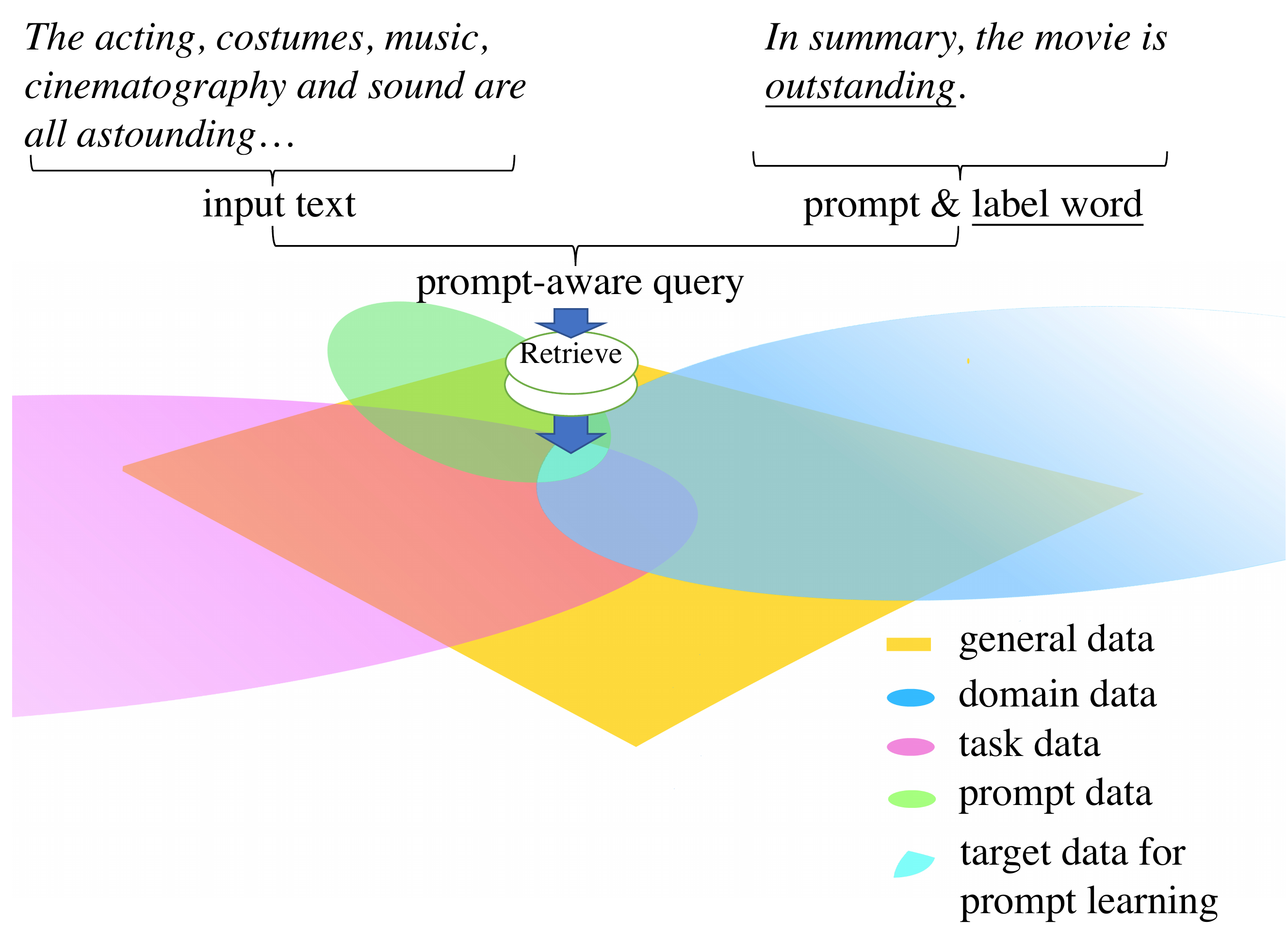}
    \caption{The distributions of data in prompt-based models. Task data, domain data, prompt data, and general data (for LM pretraining) are usually sampled from different distributions while remaining certain overlap (target data for prompt training). We aim to explore data from the overlapping area to bridge the gap between PLM and downstream tasks in prompt-based systems.}
    \label{fig:1}
\end{figure}

Prompt-based methods~\cite{Brown2020LanguageMA,liu2021pre,schick2021exploiting, li2021prefix} have received increasing attention in Natural Language Processing (NLP) recently. 
The main idea is to make the most use of pretrained language models (PLMs) by adapting an NLP task into a natural language prompt, which can then be filled by PLMs.
Take sentiment classification~\cite{socher2013recursive, 9276424} for example.
Given the sentence ``\emph{I love the movie.}'', the standard task is to make a binary classification on its sentiment polarity (i.e., \texttt{positive} or \texttt{negative}). Prompt-based methods first  transform the sentence into ``\emph{I love the movie. \underline{The movie is $\langle$\texttt{mask}$\rangle$.}}'' (the underlined text is called prompt), and then identify its polarity by checking whether PLMs tends to predict ``\emph{good}'' or ``\emph{bad}'' for the  $\langle$\texttt{mask}$\rangle$ token 
(where the predicted words are then verbalized into class labels).
The prompt-based task formulation is close to masked language modeling \cite{schick2021exploiting,schick2021s}, which is the mainstream pretraining strategy, allowing PLMs to provide rich language knowledge seamlessly.
Prompt-based methods have been shown particularly useful in zero-shot and few-shot settings \cite{petroni-etal-2019-language,yin-etal-2019-benchmarking, min2021noisy}, where with limited direct task data, prompt-based inference benefits more from large-scale pretraining than task-oriented fine-tuning.

Existing methods, however, still suffer from several potential limitations. 
First, large raw text data used for pretraining do not necessarily contain sufficient patterns that are directly related to task specific prompts~(illustrated in Figure~\ref{fig:1}). 
For instance, the prompt for a question classification task is ``\emph{\underline{Can you tell me the $\langle$\texttt{mask}$\rangle$}: What are the twin cities?}'', where $\langle$\texttt{mask}$\rangle$ should be a class label word, \emph{e.g., location, person}, etc (the correct label for this sample is \emph{definition}). However, LM pretraining data are typically \textsc{BookCorpus}~\cite{zhu2015aligning} plus \textsc{Wikipedia} corpus, where such prompts can occur scarcely in the literal or paraphrased form.
As a result, directly using PLMs to fill such handcrafted prompts across domains can lead to poor performance.
Second, to project label words to task labels, most existing work~\citep{schick2021exploiting,schick2021s,cui-etal-2021-template} uses a pre-defined verbalizer.
However, it often requires expert knowledge to build a  verbalizer that can thoroughly cover candidate words and a poorly-designed verbalizer limits the accuracy of predictions.
These problems become even more serious under zero-shot or very-few-shot settings, where prompt-based models highly rely on the generalization ability of PLMs to new tasks and domains.

We propose AdaPrompt, a framework that adapts PLMs for end tasks considering both the prompts and the verbalizer.
We are interested in addressing the above issues under a zero-shot setting, where little or no labeled training data are available for a particular task. 
The main idea is to adapt a PLM to a strong prompt-based model for an end task by exploring knowledge from its raw input data.
In particular, as shown in Figure~\ref{framework}, given a raw test set without labels, we first ask a PLM to fill a prompt template for each  input (e.g., ``{\it In summary, the movie is great.}'', where ``\emph{great}'' is filled by PLMs). Then, we use the resulting text (input text + prompt + PLM output) as a prompt-aware query to retrieve relevant data from a large unlabeled corpus.
In this manner, we can obtain a large dataset that contain both task and prompt characteristics, and we adaptively continual pretrain~\cite{gururangan2020don} the PLM on the retrieved data, which can substantially  benefit prompt-based methods on  downstream NLP tasks.


Meanwhile, we found current way of building verbalizers is also not optimal.
Given a specific task, different words can be verbalized into the same class labels. For example, a large number of adjectives can express the positive sentiment, and the best-performing candidates depend on the domain, PLM and context.
In AdaPrompt, we propose to adaptively augment verbalizers by making use of knowledge from PLMs and Natural Language Inference (NLI) models.
Take sentiment analysis for example, given \emph{``good''} and \emph{``bad''} as seed verbalizers, we first let PLMs to predict more  candidate words, such as ``\emph{amazing}'' and ``\emph{great}''.
Then, to identify if these candidates are suitable to verbalizer, we refer to an NLI model to predict whether \emph{``This movie is amazing.''} entails the meaning of \emph{``This movie is good.''}.
In this way, we can automatically expand the verbalizers.

Experiments on five text classification tasks show that AdaPrompt outperforms baseline prompt-based methods by  2.29\%-5.79\% in very-few-shot setting and 2.46\%-15.00\% in zero-shot setting on accuracy.
To our knowledge, we are the first to consider how to bridge the gap between LM pretraining and NLP downstream tasks for prompt-based NLP.
We release our code and data at \texttt{\href{https://github.com/cylnlp/AdaPrompt}{https://github.com/cylnlp/AdaPrompt}}.

\section{Related work}
\subsection{Zero/Few-shot Prompt-based NLP}
Although prompt-based methods have been used for multiple NLP tasks~\citep{Brown2020LanguageMA,raffel2020exploring, Brown2020LanguageMA,cui-etal-2021-template}, most of existing work focus on text classification~\cite{shin2020eliciting, gao2020making,min2021noisy, hu2021knowledgeable}.
A typical related work is PET~\cite{schick2021exploiting}, where \citet{schick2021exploiting} formally define \emph{pattern-verbalizer pairs} that have been widely adopted by successive works.
By using such pairs, ~\citet{schick2021exploiting,schick2021s} develop a series of work to explore the potential of PLMs, including annotating soft labels for raw training data, and data augmentation iteratively.
However, different from PET that assumes the availability of large silver training set for downstream tasks, we focus on zero and very-few-shot settings, where even unannotated task-relevant dataset is also limited~\cite{perez2021true}.
Therefore, following ~\citet{hu2021knowledgeable}, we simply focus on standard pattern-verbalizer pairs for text classification.

Prompt engineering~\cite{jiang2020can,gao2020making} focuses on how to create prompts that can better induce PLMs to make correct predictions.
Discrete prompt engineering works by replacing, deleting, inserting or paraphrasing parts of the prompt~\cite{wallace2019universal,yuan2021bartscore}.
Those methods can efficiently adapt PLMs to end tasks, but they highly reply on annotated data for tuning parameters.
Different from the above studies, we are interested in narrowing the gap between LM pretraining and NLP tasks for prompting learning in zero or very-few-shot settings.

\begin{figure*}
    \centering
    \includegraphics[width=0.95\textwidth]{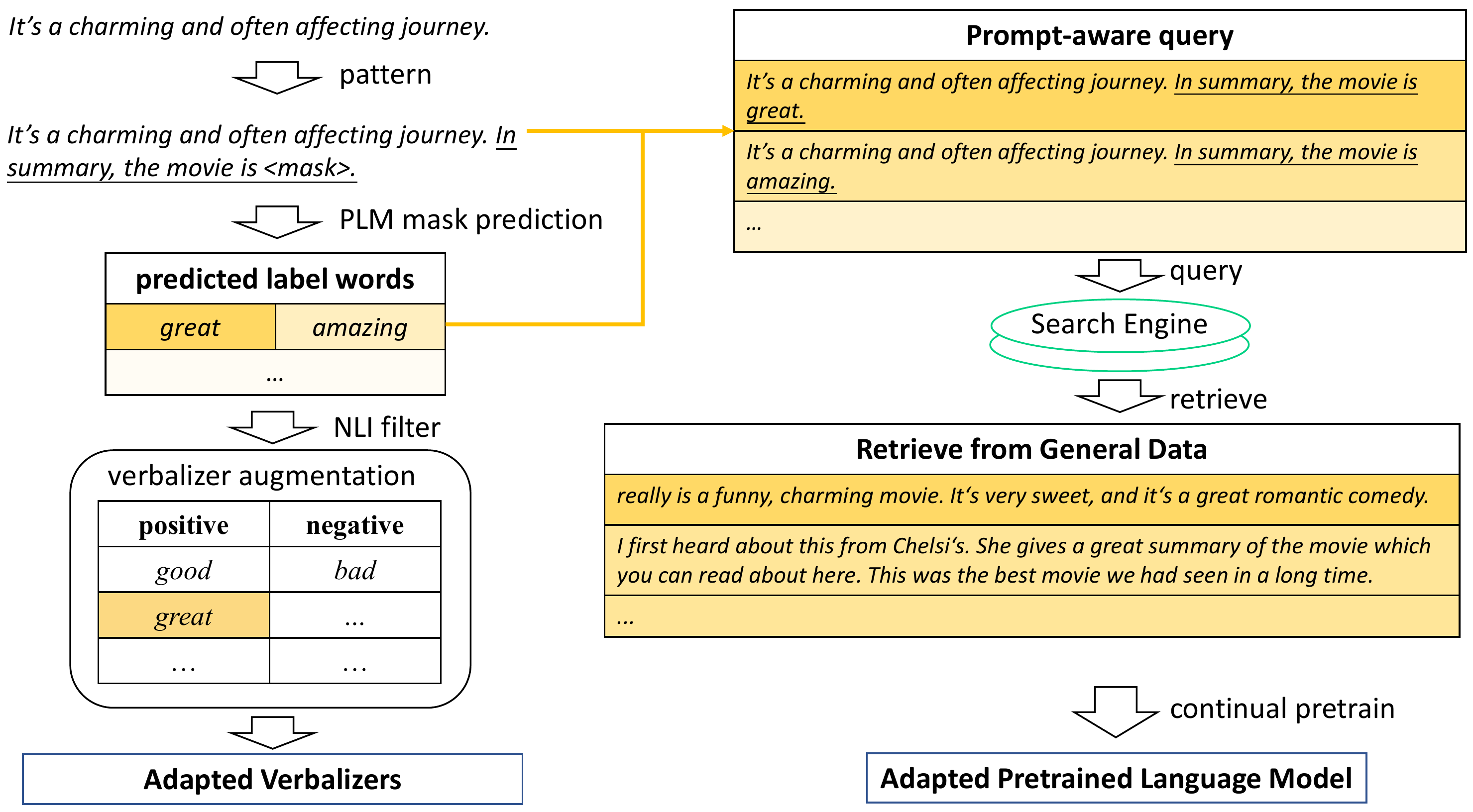}
    \caption{Overall framework of 
    AdaPrompt.}
    \label{framework}
\end{figure*}



It has been shown that using different verbalizers can also be a key factor for prompt learning~\cite{hu2021knowledgeable,cui-etal-2021-template}. 
However, manually exploring label words is time-consuming and may neglect potential candidates.
Recently, ~\citet{hu2021knowledgeable} uses multiple external knowledge bases, such as related words and sentiment dictionaries, to augment verbalizers for corresponding tasks.
Different from them, we focus on exploring knowledge in PLMs themselves. 
By making use of external NLI models AdaPrompt can select verbalizers automatically without the need of labeled task data, which is useful in zero-shot settings.

\subsection{Continual Pretraining for Domain Adaptation}

Continual pretraining~\cite{gururangan2020don} has shown benefit of optimizing a PLM to a target domain before further fine-tuning.
It can be categorised into domain adaptive  continual pretraining and task adaptive continual pretraining.
The difference is that, domain adaptive pretraining (DAPT) uses domain relevant data while task adaptive pretraining (TAPT) uses task-specific data.

Similar to continual pretraining, many recent methods highlight the merits of relying on language modeling objectives for domain adaptation.
\citet{chronopoulou-etal-2019-embarrassingly} and \citet{radford2018improving} propose to train task-specific parameters for PLMs by using an auxiliary LM loss on target domains.
Models like SciBERT~\cite{beltagy2019scibert}, DialogLM~\cite{zhong2021dialoglm}, AMRBART~\cite{bai-etal-2022-graph}, SARA-BERT~\cite{bai-etal-2022-semantic} and Dict-BERT~\cite{yu-etal-2022-dict} are PLMs that are continually pretrained on large
amounts of domain/task-specific corpora.



Data selection is a common practice in domain adaption for NLP models~\cite{moore2010intelligent,ruder2017learning,van2017dynamic}.
It has been used in machine translation~\cite{van2017dynamic,wang2018denoising}, parsing~\cite{plank2011effective,ruder2017learning} and sentiment analysis~\cite{ruder2017data}.
The main idea is to have a selection model that can distinguish in-domain and out-of-domain data.
The selection model can be a supervised classifier~\cite{aharoni2020unsupervised}, similarity-based metric~\cite{plank2011effective} or language model perplexity~\cite{moore2010intelligent}.
Very recently, \citet{yao2021nlp} propose to retrieve a small set of training data from general corpora with labeled task data as queries, finding that using LM objective on this data as an auxiliary loss can help train task-specific NLP models without pretraining.

\section{Method}
Our method is based on prompt-based text classification methods~(Section~\ref{prompt-text-classification}).
The overall procedure of AdaPrompt is shown in Figure~\ref{framework}, which can be divided into two parts: PLM  adaptation~(Section~\ref{continue-pretraining}) and verbalizer adaptation~(Section~\ref{verbalizer-adaptation}).
In Section~\ref{iterative}, we introduce a method that adapts both PLMs and verbalizers in an iterative way for continual improvements.


\subsection{Prompt-based Text Classification}\label{prompt-text-classification}

Given an input text, $\textbf{x} = (x_0, x_1,... ,x_n)$, we consider various tasks to classify the sentence into a class label $l \in \mathcal{L}$.
As mentioned in Section~\ref{intro}, the standard prompt-based method reformulates the input into a cloze-style question and identifies its label by checking PLMs' predictions.
Table~1 shows the prompt templates and verbalizer patterns for the SST-2 \cite{socher2013recursive}, Yelp \cite{zhang2015character}, AGNews \cite{zhang2015character}, TREC \cite{voorhees2000building} and  DBPedia \cite{lehmann2015dbpedia} datasets, which cover sentiment classification, topic classification and question classification tasks.
Formally, let $\mathcal{M}$ be a language model pretrained on large-scale general data, and $\langle$\texttt{mask}$\rangle$ be the mask token. 
The prompt-based method first defines a \emph{pattern} function, $Prompt$, that converts \textbf{x} into a cloze-style question containing $\langle$\texttt{mask}$\rangle$. 
Then, it defines a \emph{verbalizer} function $v$, which maps a small set of pre-defined verbalizer words ($\mathcal{Y}$) predicted at the position of \texttt{<mask>} into class labels, i.e., $v: \mathcal{Y} \mapsto \mathcal{L}$.

Take sentiment classification for movie review for instance.
The task is to classify the sentiment polarity, where $\mathcal{L} = \{\texttt{positive, negative}\}$. For an input \textbf{x}, we choose the pattern:

{$\quad Prompt = $``\textbf{x}. \emph{In summary, the movie is $\langle$\texttt{mask}$\rangle$}.''}

Then we define a verbalizer that maps $\mathcal{Y} = \{\emph{``good'', ``bad''}\}$ into $\mathcal{L}$:

{\quad $v(\emph{``good''}) = \texttt{positive} $};

{\quad $v(\emph{``bad''}) = \texttt{negative} $}

Given an example:

{$\quad \textbf{x} =  $ ``\emph{It's a charming journey.''}},

we can convert the input into a cloze-style question using $Prompt$:

{$\quad Prompt(\textbf{x}) =  $ ``\emph{It's a charming journey. In summary, the movie is $\langle$\texttt{mask}$\rangle$.''}}

Using such \emph{pattern-verbalizer} pairs,
we ask $\mathcal{M}$ to directly give scores $s$ for each label $l \in\mathcal{L}$ as:
\begin{equation}\label{zero-shot}
    s(l|\textbf{x}) = Pr[\texttt{<mask>}=y|Prompt(\textbf{x}),\mathcal{M}]\\
\end{equation}
where $l = v(y)$.
The predicted label is:
\begin{equation}\label{argmax}
    \hat{l} = \mathop{\arg\max}\limits_{l \in \mathcal{L}} s(l|\textbf{x})
\end{equation}

\subsection{Adaptively Retrieve Data for Continual Pretraining}\label{continue-pretraining}

As discussed in the Section~\ref{intro}, the lack of domain adaptation can be a potential challenge for prompt-based NLP models, especially under zero-shot and very-few-shot settings. 
To tackle this problem, we propose to build a continual pretraining dataset by retrieving from general corpora, with unannoated test texts, designed prompts and label words as queries.
In this way, we can obtain task-relevant data for any tasks or domains, using only test input.
Meanwhile, prompt and verbalzier information is also considered during the retrieval process, leading to a more comprehensive dataset for prompt-aware continual pretraining. 

Formally, given a retrieval query $q$, a retrieval engine $\mathcal{E}_D$ indexed on a large general dataset $\mathcal{D}$ can return a set of similar text $d_{q} = \mathcal{E}_D(q)$.
To obtain prompt-aware data that can not only adapt PLMs to target domains but also make PLMs more sensitive to prompts, we include both task and prompt characteristics when building queries.
As shown in Figure~\ref{framework}, for a raw input text $\textbf{x}$ in text data, we first convert it into $Prompt(\textbf{x})$, and obtain a set of predicted label words using a PLM $\mathcal{M}$:
\begin{equation}\label{prediction}
    \mathcal{O} = \mathcal{M}(Prompt(\textbf{x}))
\end{equation}
where $\mathcal{O} = \{o_1, o_2, ..., o_{|\mathcal{O}|}\}$ are the top-$|\mathcal{O}|$ predictions.
We replace the mask token in $P(\textbf{x})$ with $o_i$, to form a list $Q$ of queries. For example:
\begin{equation}
    Q = \{q_1,... ,q_{|\mathcal{O}|}\},
\end{equation}

where $q_i = ``\textbf{x}.\ \emph{In summary, the movie is}\ o_i."$

With this  set of prompt-based queries, we  retrieve prompt-aware data $\mathcal{D}_p$, which is a small subset of the  general data.
In this work, we use ElasticSearch\footnote{\href{https://www.elastic.co}{https://www.elastic.co}} indexed on a large general corpus as the search engine and we ask it to return a list of top-$k$ texts that match the query.
As shown in Figure~\ref{framework}, one test input can lead to multiple prompt-aware queries because the masked token in the prompt can be replaced by the $|\mathcal{O}|$ predictions.
In addition, given one query, ElasticSearch can also give multiple returns with demanded $k$.

We continue to pretrain the PLM $\mathcal{M}$ on $\mathcal{D}_p$ with masked language modeling loss and obtain an  adapted PLM $\mathcal{M}_{\mathcal{D}_p}$.
$\mathcal{M}_{\mathcal{D}_p}$  now contains richer knowledge of both the target domain and the prompts. It can be used to replace $\mathcal{M}$ in Eq.~\ref{zero-shot} for zero-shot text classification.

\begin{algorithm}[tb]
\small
  \caption{Verbalizer Adaptation}
  \label{vaalgo}
\begin{algorithmic}
  \STATE {\bfseries Input:} prompt $P$, seed verbalizer words $y \in \mathcal{Y}_l$, candidate words $c \in \mathcal{C}$ and an NLI system $\mathcal{N}$
  \FOR{$c$ {\bfseries in} $\mathcal{C}$} 
  \IF{$\mathcal{N}(f(P, y), fill(P, c))=Entail$ 
  \\\quad or $\mathcal{N}(fill(P, c), f(P, y))=Entail$} 
  \STATE add $c$ in $\mathcal{Y}_l$
  \ENDIF
  \ENDFOR
  \STATE Return $\mathcal{Y}_l$
\end{algorithmic}
\end{algorithm}

\subsection{Iterative Adaptation}\label{iterative}
After obtaining $\mathcal{M}_{\mathcal{D}_p}$, we can iterate the process by replacing $\mathcal{M}$ with $\mathcal{M}_{\mathcal{D}_p}$ in Eq.~\ref{prediction}, and obtain an iterative set of predicted words and a list of queries marked as $\mathcal{O}^\prime$ and $Q^\prime$.
Given that $\mathcal{O}^\prime$ contains more in-domain knowledge, we can retrieve higher quality pretraining data with more task relevant information, using $Q^\prime$ to query the $\mathcal{E}_D$.
In this way, we obtain a new version of  $\mathcal{D}_p^\prime$, and a new continual pretrained PLM $\mathcal{M}_{\mathcal{D}_p}^\prime$, which can also be used for zero-shot predictions using Eq.~\ref{zero-shot}.
In this work, we conduct this procedure twice.

\subsection{Adaptive Verbalizer Augmentation}\label{verbalizer-adaptation}
As described in Section~\ref{prompt-text-classification}, the regular prompt-based method defines the \emph{verbalizer} that maps predicted label word into  task classes, such as ``\emph{good}'' for \texttt{positive} and ``\emph{bad}'' for \texttt{negative}.
However,  predefined verbalizer can be limited.
To expand this verbalizer, we first 
infer top-$|\mathcal{O}|$ label words at mask token position over all inputs in test set.
We  filter the predicted words and obtain a set of high frequent  words $\mathcal{C}$ as candidates for verbalizer augmentation.
Then, we propose a new method for exploring useful verbalizer words by using knowledge from a  Natural Language Entailment model.


Specifically, given a seed verbalizer word $y_l \in \mathcal{Y}_l$ for label $l$, and a candidate word $c \in \mathcal{C}$, we compare whether a prompt filled by $y_l $ is entailed with the prompt filled by $c$. 
The pseudo code is shown in Algorithm~\ref{vaalgo}.
If entailment relation holds for this pair, we add $c$ add to $\mathcal{Y}_l$.
And the new $\mathcal{Y}$ which can be considered as an augmented verbalizer.

After obtaining the augmented set of verbalizer words,
Eq.~\ref{zero-shot} can be rewritten as:
\begin{equation}\label{zero-shot2}
    s(l|\textbf{x}) = \frac{1}{|\mathcal{Y}_l|} \sum_{y \in \mathcal{Y}_l}{Pr[\texttt{<mask>}=y|Prompt(\textbf{x}),\mathcal{M}]}
\end{equation}
and we can still use Eq.~\ref{argmax} for prediction.



\begin{table*}[t]
\setlength\tabcolsep{5pt}
\small
\centering
{
\begin{tabular}{l|c|c|c|c}
\hline
\textbf{Dataset} &\textbf{Class}&\textbf{Objective}& \textbf{Prompt Template} & \textbf{Verbalizer}\\
\hline
SST-2 & 2& sentiment& \makecell[c]{\texttt{Text} \emph{In summary, this movie is $\langle$mask$\rangle$.}} & \emph{``good''}, \emph{``bad''}\\
\hline
Yelp & 2& sentiment & \makecell[c]{\texttt{Text}  \emph{In summary, this restaurant is $\langle$mask$\rangle$.}} & \emph{``good''}, \emph{``bad''}\\
\hline
AGNews &4& news topic & \emph{[Category: $\langle$mask$\rangle$]} \texttt{Title} , \texttt{Body}  & \emph{``Sport''}, \emph{``Tech''}, \emph{``Business''}, \emph{``World''}\\
\hline
TREC &6& question & \emph{Can you tell me the $\langle$mask$\rangle$} \texttt{Text}  & \makecell[c]{\emph{``explanation''}, \emph{``description''}, \emph{``person''},\\ \emph{``location''}, \emph{``number''}, \emph{``entity''}}\\
\hline
DBPedia &14& ontology & \emph{Introduction to the $\langle$mask$\rangle$} \texttt{Text}  & \makecell[c]{\emph{``company''}, \emph{``school''}, \emph{``artist''},\emph{``film''},\\ \emph{``book''}, \emph{``plan''}, \emph{``building''}, \emph{``village''},\\ \emph{``animal''}, \emph{``sport''},  \emph{``album''},\\ \emph{``officer''}, \emph{``scenery''}, \emph{``transportation''}} \\
\hline
\end{tabular}
}
\caption{Datasets used in this paper with seed prompts and verbalizer words. 
Each seed verbalizer word corresponds to a class label.
}
\label{prompts-examples}
\end{table*}

\begin{table}[!t]
\centering
\setlength\tabcolsep{3.5pt}
\small
{
\begin{tabular}{l|cccc}
\hline
\textbf{Dataset}& \textbf{test set} & \textbf{Top-$|\mathcal{O}|$} & \textbf{$E_{space}$} &\textbf{Resulting Data}\\
\hline
TREC & 500 & 20 & 100& 60k \\
SST-2 & 872 & 20 & 100 & 205k \\
AGNews & 7,600 & 10 & 50 & 414k \\
YELP & 38,000 & 1 & 50 & 267k\\
DBPedia & 70,000 & 1 & 50 &  1,301k\\
\hline
\end{tabular}
}
\caption{Data statistics for datasets. 
$E_{space}$ corresponds to the ElasticSearch space.
Note that the resulting data size is calculated after data de-duplication.}
\label{test-set-stat}
\end{table}

\section{Experiments}
\subsection{Datasets and Prompts}
To evaluate our methods, we conduct experiments on five benchmarks: SST-2 \cite{socher2013recursive}, Yelp \cite{zhang2015character}, AGNews \cite{zhang2015character}, TREC \cite{voorhees2000building} and  DBPedia \cite{lehmann2015dbpedia} datasets.
Table~\ref{prompts-examples} shows prompt templates and seed verbalizer words that we use for each dataset.
For AGNews and YELP, we adapt patterns and verbalizers from PET~\cite{schick2021exploiting} since it is the basic prompt-based method that has been mostly widely used.

\textbf{AGNews} is a text classification dataset in the domain of News.
Given a headline and a main text body, the model is require to classify the news into one of the classes: (1) World, (2) Sports, (3) Business or (4) Science/Tech.

\textbf{YELP} is a sentiment analysis dataset. 
Given a restaurant review, the task is to predict whether the review is positive or negative.

\textbf{SST-2} is a sentiment analysis dataset similar to YELP but  its domain is movie reviews.
Thus, we use the same seed prompt and verbalizer words as for YELP, but change ``\emph{restaurant}'' in prompt template to ``\emph{movie}''.

\textbf{DBPedia 2014} is an ontology classification dataset, extracted from DBPedia 2014 with 14 non-overlap classes, such as Educational Institution and Office Holder. 
We define two patterns for this task:

{$\quad P1(\textbf{x}) =  $ ``\emph{Description to the $\langle$\texttt{mask}$\rangle$ \textbf{x}''}}

{$\quad P2(\textbf{x}) =  $ ``\emph{Introduction to the $\langle$\texttt{mask}$\rangle$ \textbf{x}''}}

and we use $P2$ as the seed pattern.

\textbf{TREC-10} is a question classification dataset. 
Given a question, the task is identify the objective that the question asks, and classify it into one of six classes, such as a definition question or a numeric question.
We define two patterns for this task:

{$\quad P1(\textbf{x}) =  $ ``\emph{Tell me the $\langle$\texttt{mask}$\rangle$ \textbf{x}''}}

{$\quad P2(\textbf{x}) =  $ ``\emph{Can you tell me the $\langle$\texttt{mask}$\rangle$: \textbf{x}''}}

and $P2$ as the seed prompt.

\begin{table*}[htbp]
    \setlength\tabcolsep{3pt}

    \small
    \centering
    \begin{tabular}{l|ccccc|c}
    \hline
    \textbf{Models} &\textbf{SST-2} & \textbf{Yelp} &\textbf{AGNEWS} &\textbf{DBPedia} &\textbf{TREC} &\textbf{Avg.}\\
    \hline
    
    GPT-2 & $63.00/\ \ \ \ \ NA (NA)$ & $--$ & $59.80/ \ \ \ NA(NA)$ & $32.30/ \ \ \ NA(NA)$  & $38.70/ \ \ \ NA(NA)$ & $--$\\
    Channel& $77.10/\ \ \ \ \ NA (NA)$ & $--$ & $61.80/ \ \ \ NA(NA)$ & $51.40/\ \ \ NA(NA)$ & $30.50/\ \ \ NA(NA)$ &$--$\\
    GPT-3& $75.80/\ \ 0.00(75.80)$ & $--$ & $73.90/0.00(73.90)$ & $59.70/0.00(59.70)$ & $57.40/0.00(57.40)$ &$--$\\
    \textsc{R.}& $64.56/16.77(88.99)$ & $72.63/ \ \ 6.34(87.97)$ & $69.52/ 6.96(78.76)$ & $56.32/0.49(56.67)$ & $45.50/0.14(45.60)$ &$61.71$\\
    \hline
    Ada & $75.92/17.36(91.28)$& $75.09/17.57(89.25)$& $76.55/7.28(84.95)$ &$70.95/8.80(77.17)$ &$60.50/3.54(63.00)$ & $71.80$\\
    \hline
    iAda & $77.18/17.96(91.74)$  & $75.81/18.05(90.41)$& $74.28/9.00(83.37)$ &$73.01/6.70(77.92)$&$61.10/1.27(62.00)$ & $72.28$\\
    \hline
    \end{tabular}
    
    \caption{Zero-shot results. We report average accuracy and standard deviation of different patterns here. Results of the best patterns are shown in brackets.
    The Avg. reports the overall averaged results.
    \textsc{R.} stands for \textsc{RoBERTa}-large.
    Ada and iAda denote to AdaPrompt and iterative AdaPrompt based on \textsc{RoBERTa}-large, respectively.
    The results of GPT-2 large and Channel are from~\cite{min2021noisy}, and Channel is based on GPT-2 large.
    GPT-3 results are reported by \citet{zhao2021calibrate}, using GPT-3 ($175B$).
    $NA$ denotes to that results are not reported.
    For GPT-3~\cite{zhao2021calibrate}, they only use a fixed prompt format.}
    \label{tab:main_results}
\end{table*}
\subsection{Settings}\label{settings}
In this work, we take \textsc{RoBERTa}-large~\cite{liu2019RoBERTa} as our foundation PLM and adopt pattern-verbalizer pairs from ~\cite{schick2021exploiting} (Section~\ref{prompt-text-classification}) as the baseline setting which is widely used and can be easily extended to other methods~\cite{shin2020eliciting}.

We conduct experiments in zero-shot and few-shot settings.
In the zero-shot setting, we directly use PLMs to infer label words at masked positions.
Under the few-shot setting, we follow~\citet{schick2021exploiting} and~\citet{hu2021knowledgeable} and use prompt-tuning, which directly fine-tunes a LM given a small set of annotated data and prompts.


For zero-shot settings, the choice of hyper-parameters is based on previous work~\cite{gao2020making,schick2021exploiting,schick2021s}.
For all continual pretraining, we use a learning rate of $1e^{-5}$, batch size of 96.
We train each model for 3 epochs and use the checkpoint at 500  steps for evaluation.

For few-shot settings, we evaluate our models 
with 10, 50, 100 training samples.
We follow previous work~\citep{hu2021knowledgeable,schick2021exploiting,gao2020making} and repeat the training and evaluation for 5 times using different seed, and report the averaged scores for each datasets.

\begin{table*}[t]
\small
\centering
{
\begin{tabular}{l|c|ccccc|c}
\hline
\textbf{$|\emph{T}|$} &\textbf{Models}&\textbf{SST-2} & \textbf{Yelp} &\textbf{AGNEWS} &\textbf{DBPedia} &\textbf{TREC}&\textbf{Avg.}\\
\hline
\multirow{2}{*}{10} & \textsc{RoBERTa} & $84.97\pm9.88 $ & $86.84\pm16.08$ & $78.42\pm6.23$ & $86.78\pm1.10$ & $45.56\pm9.55$ & $76.51$ \\
& AdaPrompt & $90.42\pm 1.63 $ & $89.13\pm13.30$ & $84.21\pm2.00$ & $91.68\pm1.84$ & $57.56\pm7.85$& $82.60$ \\
\hline
\multirow{2}{*}{50} &  \textsc{RoBERTa} &$92.56\pm1.31$&  $95.87\pm \ \ 0.57$& $85.50\pm1.36$ & $94.72\pm 0.49$ & $73.88\pm 3.13$ &  $88.51$\\
& AdaPrompt & $92.75\pm1.03 $ & $95.74\pm \ \ 0.89$ & $86.29\pm0.80$ &  $94.59\pm0.71$  & $78.42\pm6.17$ & $89.56$\\
\hline
\multirow{2}{*}{100} & \textsc{RoBERTa} & $92.40\pm1.04$ & $95.89\pm \ \ 0.68$ & $87.29\pm1.31$ & $95.59\pm0.52$ & $86.30\pm2.14$ & $91.49$ \\
& AdaPrompt & $92.75\pm 0.68$ & $95.93\pm \ \ 0.95$ & $87.98\pm0.65$ &  $95.60\pm0.51$  & $87.58\pm1.38$ & $91.97$\\
\hline
\end{tabular}
}
\caption{Average accuracy and standard deviation on SST-2, YELP, AGNews, DBPedia and TREC under few-shot settings. 
$|\emph{T}|$ is the training set size.
Each experiment is repeated $5$ times using different seeds.
}
\label{few-results}
\end{table*}

\paragraph{Prompt-Aware Data Retrieval}
We  take pretrain data of the \textsc{RoBERTa} model (  \textsc{BookCorpus}~\cite{zhu2015aligning}, \textsc{Wikipedia}, \textsc{Ccnews}~\cite{ccnews}, \textsc{Stories}~\cite{trinh2018simple}, and \textsc{OpenWebText}~\cite{openweb}) as the general dataset to query from.
We  index them on sentence level with ElasticSearch and consider TF-IDF as the similarity metric.

Table~\ref{test-set-stat} presents the statistics of evaluation datasets used in this paper.
TREC and SST contain smaller test sets, while YELP and DBPedia contain much larger test sets.
To balance the retrieved data size, we set different top-$|\mathcal{O}|$ for predicted words and ElasticSearch space ($k$) for different datasets based on our practical experience.
In other words, given one test input, we have $|\mathcal{O}| \times k$ data.
After de-duplication, the resulting retrieved data sizes are shown in Table~\ref{test-set-stat}.


\paragraph{Verbalizer Augmentation}
To obtain possible verbalizers that can better represent classes, we first obtain top-$N$ predicted words given a test sample ($N=20$ for SST-2 and TREC,  $N=10$ for AGNews and $N=5$ for YELP and DBPedia, considering their test set sizes).
We set the number of candidate words $|\mathcal{C}|$ = $20\times|\mathcal{L}|$, where $|\mathcal{L}|$ is number of classes.
We use a \textsc{RoBERTa}-large model fine-tuned on MNLI~\cite{N18-1101}, as the entailment model for identifying potential verbalizer words for augmentation.
Candidate with probability higher than a threshold $t$ is then added to the augmented verbalizer. We set $t=0.4$ by experiments.

For comparison, we also use Word2Vec
\cite{mikolov2013efficient} to obtain word vectors and explore potential verbalizer words by their similarity with the seed verbalizer words. 

\subsection{Results}
\subsubsection{Main Results}

\begin{table*}
    \setlength\tabcolsep{3pt}
    \small
    \centering
    \begin{tabular}{l|ccccc|c}
    \hline
    \textbf{Models} &\textbf{SST-2} & \textbf{Yelp} &\textbf{AGNEWS} &\textbf{DBPedia} &\textbf{TREC} &\textbf{Avg.}\\
    \hline
    AdaPrompt & $75.92\pm17.36$& $75.09\pm17.57$& $76.55\pm07.28$ &$70.95\pm08.80$ & $60.50\pm03.54$& $71.80$\\
    \hline
    \texttt{-va} &$71.07\pm13.58$& $71.04\pm15.57$& $72.16\pm05.78$ &$65.90\pm02.71$&$45.40\pm01.13$& $65.11$\\ 
    \texttt{-CP} &$72.16\pm16.35$& $75.72\pm17.79$& $75.70\pm07.88$ & $50.95\pm00.09$ & $58.70\pm 03.25$  & $66.65$ \\
    \texttt{-PR} &$71.22\pm15.55$& $74.85\pm17.51$& $75.12\pm05.71$ & $70.40\pm07.48$ & $58.60\pm00.57$ & $70.04$
    \\
    \texttt{-CP-va} &$64.56\pm16.77$& $72.63\pm16.34$ & $69.52\pm06.96$ & $56.32\pm00.49$ &$45.50\pm00.14$& $61.71$\\
    \hline

    \end{tabular}
    
    \caption{
    Experimental results of ablation study.
    ``\texttt{-}'' means ``without'' here. 
    \texttt{va}: verbalizer augmentation, \texttt{CP}: Continual Pretraining, \texttt{PR}: Prompt-aware Retrieval.
    Note that \texttt{-PR} means we do not use prompt-aware retrieval, but simply use raw test input data for retrieval and continual pretraining, refered as \emph{in-domain adaptation.}}
    
    \label{tab:ablation}
\end{table*}

\begin{table}
\centering
\setlength\tabcolsep{3.5pt}
\small
{
\begin{tabular}{l|c|c}
\hline
\textbf{Model}&\textbf{SST-2} &\textbf{DBPedia}\\
\hline
\textsc{RoBERTa} & $64.82\pm11.62$ & $56.49\pm00.41$\\
AdaPrompt & $73.05\pm13.08$ & $70.97\pm08.87$ \\
\hline
\end{tabular}
}
\caption{
Model performance tested on \emph{unseen} test set.
We report averaged accuracy and standard deviation.
}
\label{generalization}
\end{table}

\begin{table}[]\small \centering
\begin{tabular}{l|cccc}
\hline
\multicolumn{5}{c}{\textbf{SST-2}}                                    \\ \hline
\multicolumn{1}{l|}{ $E_{space}$} & $1$& $10$ & $50$& $100$ \\\hline
\multicolumn{1}{l|}{Size} & 3k & 23k & 98k & 205k   \\ \hline
\multicolumn{1}{l|}{\multirow{2}{*}{Accuracy}}&$73.54$&$75.06$&$75.95$&$75.92$ \\
\multicolumn{1}{l|}{}   &$\pm16.77$&$\pm17.34$&$\pm17.73$&$\pm17.36$  \\ \hline
\multicolumn{5}{c}{\textbf{DBPedia}}                                      \\ \hline
\multicolumn{1}{l|}{$E_{space}$} & 1& 5 & 25& 50\\\hline
\multicolumn{1}{l|}{Size} & 58k & 235k & 708k & 1,301k  \\\hline
\multicolumn{1}{l|}{\multirow{2}{*}{Accuracy}} &$70.64$&$71.39$&$74.13$&$70.95$  \\
\multicolumn{1}{l|}{}  &$\pm 9.66$&$\pm10.78$&$\pm7.51$ &$\pm8.80$
 \\ \hline
\end{tabular}
    \caption{Analysis on retrieved data size.
    Data sizes are calculated after de-duplication.}
    \label{tab:elastic-engine-space}
\end{table}

\paragraph{Zero-shot Performance}\label{setting1}
In zero-shot setting, we compare AdaPrompt with prompt-based methods using \textsc{RoBERTa}~\cite{schick2021exploiting}, GPT-2~\cite{gao2020making} and GPT-3~\cite{zhao2021calibrate}, respectively.
The Channel refers to noisy channel model~\cite{min2021noisy} based on GPT-2.
Table~\ref{tab:main_results} presents the results under zero-shot setting.
Following previous work~\cite{schick2021exploiting, schick2021s}, we report average accuracy, standard deviation and accuracy of the best pattern over different patterns.

First, compared with our foundation model, \textsc{RoBERTa}-large, we see that AdaPrompt consistently outperforms regular prompt-based methods on all datasets with better average performance and best pattern performance, bringing a $2.46\sim14.63$ improvement.
It is noticeable that  AdaPrompt outperforms GPT-3 in zero-shot setting, which is a huge model with 175$B$ parameters pretrained on a gigantic corpus.
This confirms the effectiveness of AdaPrompt in domain adaptation.
We observe that iterative AdaPrompt can further bring improvements on most datasets (SST-2, YELP and DBPedia).
This directly demonstrates that PLMs continual pretrained on the retrieved data can be more adaptive to downstream tasks, and thus generate more task relevant label words, which can serve as a source to find better texts.
Performance of iterative AdaPrompt (iAda) decreases on AGNEWS, we believe this is because this news dataset is similar with general data used for pretraining \textsc{RoBERTa}, and thus continual pretraining on such retrieved data can be less useful.
Finally, we see that AdaPrompt improves over 10.09 accuracy of the overall performance.

\paragraph{Few-shot Performance}\label{setting2}
Table~\ref{few-results} reports the experimental results in few shot setting.
Each experiment is repeated $5$ times using different seeds and we report the average accuracy and standard deviation.
To explore whether AdaPrompt can consistently bring improvement to \textsc{RoBERTa}, we conduct experiments using 10, 50, 100 samples, respectively.

Compared with \textsc{RoBERTa}-large baseline, under few-shot setting, AdaPrompt can still improve model performance.
Although the relative improvement decreases as the size of training set improves, we can see that AdaPrompt outperforms \textsc{RoBERTa} over all tasks in all few-shot settings.
In particular, AdaPrompt outperforms standard \textsc{RoBERTa} models by $2.29\sim5.79\%$ in $10$-shot setting, showing that it is useful in the very-few-shot setting.


\subsubsection{Ablation Study}~\label{ablation}
To study the effectiveness of continual pretraining on prompt-aware data and verbalier augmentation, we conduct ablation experiments by removing continual pretraining (\texttt{CP}) or verbalizer augmentation (\texttt{va}).
As shown in Table~\ref{tab:ablation}, We can see that compared with foundation model (\texttt{-CP-va}, $61.71$ acc. on average), continual pretraining and verbalizer augmentation can both bring improvement to model performance ($5.31$ and $5.89$ acc. on average, respectively), and the model has the best results when two methods are combined together (AdaPrompt), suggesting these two methods can benefit each other.

In addition, we investigate the influence on model performance by removing  prompt-aware retrieval and only retrieving with raw texts.
From the table we can see that on all datasets, using prompt-augmented queries (AdaPrompt) give substantially stronger results. Take SST-2  for example, the accuracy is $71.22$ (SST-2 \texttt{-PR}) given only raw input queries, but $75.92$ with prompt-augmented queries, with a $4.7$ absolute improvement. This shows that continual pretraining using prompt-aware data is highly beneficial to zero-shot prompt-based NLP.


\subsection{Analysis}

\begin{table*}[t]\small \centering
\begin{tabular}{l|ccccc|c}
\hline
\textbf{Dataset} & \textbf{SST-2}   &   \textbf{YELP}     &   \textbf{AGNEWS}     & \textbf{DBPedia} &\textbf{TREC} &\textbf{Avg.}\\ \hline
$va_w$  &    $74.91\pm11.71$   &   $75.39\pm17.47$    &   $69.07\pm06.70$    &   $55.32\pm11.33$    &   $60.60\pm03.39$   & $67.06$\\
$va_m$    &   $75.92\pm17.36$ &   $75.09\pm17.57$    &   $76.55\pm07.28$      &    $70.95\pm08.80$    & $60.50\pm03.54$   &$71.80$   \\
$va_{k}$    & $69.07\pm15.80$    &   $74.64\pm17.55$    &   $60.15\pm07.79$     &   
$74.85\pm17.50$ &   $24.00\pm00.57$ & $60.54$\\ \hline
\end{tabular}
\caption{Model performance of
AdaPrompt using different verbalizer augmentation strategies.
$va_w$: using word2vec similarity.
$va_m$: using \textsc{RoBERTa} trained on MNLI.
$va_{k}$: using most related words/sentiment dictionary.
Avg. refers to overall averaged results.
}
\label{va}
\end{table*}

\paragraph{Generalization Capability}

For experiments in section~\ref{setting1}, we use task test set as the sources to build queries for retrieving pretraining data.
However, in a more general setting, we want to learn when the query data and test set are different, whether AdaPrompt can still generalize to this test set.
To this end, we build an  {\it unseen} test set by using the original training set of SST-2 and DBPedia. We then evaluate models (trained using  queries from the origin test set) on this unseen test set.  
As shown in Table~\ref{generalization}, AdaPrompt achieves $73.05$ and $70.97$ accuracy on SST-2 and DBPedia, respectively. 
Compared with performance on original test set  (Table~\ref{tab:main_results}), although the performance of AdaPrompt sightly decreases when evaluated on  SST-2 unseen test set, it can still outperform \textsc{RoBERTa} by a large margin ($+8.23$).
It demonstrates that AdaPrompt has a strong generalization ability when query data and test set are different.


\paragraph{Size of Retrieved Data }
As stated, Elasticsearch returns top-$k$ texts in the order of matching scores.
Using a smaller $k$, the retrieved data are more textual related to the query, while using a larger $k$, the retrieved data can  contain certain noise.
To compare the effects of different sizes of retrieved data for continual pretraining, 
We set $k$ to 1, 10, 50  100 for the SST-2 and set $k$ to 1, 5, 25, 50 for DBPedia, respectively.
As shown in Table~\ref{tab:elastic-engine-space}, we see that accuracy rises in the beginning when retrieval size increases. 
But as the retrieval size grows bigger, the accuracy starts to decrease slightly. 
This can be explained by that the lower-ranked retrieved data have a lower relevance to the target task, which introduces more noise in continual pretraining. 
We use fixed $k$  for our experiments in zero-shot settings (Section~\ref{settings}), due to lack of a validation set. In few-shot settings, in practice,  $k$  can be considered as a hyperparameter and tuned over validation data.





\paragraph{The Effect of Verbalizer Strategies}
Table~\ref{va} compares the model performance when using different verbalizer augmentation strategies, namely using NLI model and word similarity~(Section~\ref{settings}).
Additional, we compare AdaPrompt with a verbalizer augmentation method using knowledge base (KB)~\cite{hu2021knowledgeable}~\footnote{For sentiment analysis tasks, we take sentiment words shown in~\cite{hu2021knowledgeable}, which are adopted from \url{https://www.enchantedlearning.com/wordlist/}; for other tasks, we use most related words: \url{https://relatedwords.org/}.}.
To set a fair comparison, we limit the verbalizer word set for each label within $5$.
We report average accuracy and standard deviation here.

Results show that, compared with using word similarity to select candidate words and directly using KBs to augment verbalizer words, using NLI to augment verbalizer words gives  better performance on most tasks, and is also more stable.
We also find that using KBs to augment verbalizer words gives better performance on the DBPedia tasks, but much worse performance on the TREC task.
This can be because TREC is less close to topic classification~\cite{min2021noisy}, and directly using the most related words can be noisy.
This also suggests that more sophisticated strategy that cares of tasks and prompt information can be useful, which we leave for future work.

\begin{table}
\centering
\setlength\tabcolsep{3.5pt}
\small
{
\begin{tabular}{l|c|c}
\hline
\textbf{Model}&\textbf{Size} &\textbf{SST-2}\\
\hline

Albert & $17M$ & $54.67\pm \ \ 3.30(58.94)$\\
Albert+AdaPrompt & $17M$ & $58.51\pm\ \ 5.79(63.99)$ \\
\hline
Bert & $340M$ & $58.03\pm \ \ 6.18(63.53)$\\
Bert+AdaPrompt & $340M$ & $68.89\pm16.11(85.67)$ \\
\hline
\textsc{RoBERTa}& $355M$  & $64.56\pm16.77(88.99)$\\
\textsc{RoBERTa}+AdaPrompt & $355M$ & $77.18\pm17.96(91.74)$ \\
\hline
\end{tabular}
}
\caption{
We report average accuracy and standard deviation here. Results of best patterns are shown in the bracket.
}
\label{bert-albert-RoBERTa}
\end{table}

\paragraph{AdaPrompt with different PLMs}~\label{different_PLM}
We apply AdaPrompt with different PLMs (Bert-large, Albert-large and \textsc{RoBERTa}-large). We report  experimental results on the SST-2 dataset in
Table~\ref{bert-albert-RoBERTa}.
Although the performance of different models varies, we observe that AdaPrompt can consistently bring huge improvement over all models.
We also find that model performance increases with model size.
AdaPrompt using \textsc{RoBERTa}-large outperforms other models overall performance by a large margin ($8.29\sim18.67$) and achieves $91.74$ accuracy with the best pattern.

\section{Conclusion}
We investigated AdaPrompt, a zero-shot prompt-based method for NLP that makes use of test input data and prompts for adaptive continual pretraining and verbalizer selection. Results on five classification datasets show that AdaPrompt improves over a standard prompt method by large margins. In particular, retrieving relevant data for continual pretraining of a language model can serve to warm-up the model for both domain adaptation and prompt-filling tasks. In addition, an NLI model allows effective selection of filled tokens to achieve improved performance.

\section*{Limitation}
We acknowledge two major limitations
of this work:
\begin{enumerate}
    \item We only tested AdaPrompt on text classification tasks. The intention is to use this clear setting to compare with other prompt-based models. However, it is possible to extend AdaPrompt to other natural language 
understanding tasks or languages, which we leave for future exploration.
\item We only tested with ElasticSearch as the search method. However, there are signals showing the quality of retrieved text is constrained to the search engines. A better configuration or model of the search method might further improve AdaPrompt.
\end{enumerate}


\section*{Acknowledgements}
Yue Zhang is the corresponding author.
We appreciate all reviewers for their comments.


\bibliography{anthology,custom}
\bibliographystyle{acl_natbib}

\clearpage
\appendix

\end{document}